\documentclass{article}
\usepackage{spconf,amsmath,graphicx}
\usepackage[inkscapelatex=false]{svg}

\usepackage{subcaption}

\usepackage{url}
\usepackage{enumitem}
\setlist{nosep, leftmargin=14pt}

\usepackage{mwe} 

\usepackage{algorithm}
\usepackage{algpseudocode}
\usepackage{amsmath, amssymb}

\title{MRIQT: Physics-Aware Diffusion Model for Image Quality Transfer in Neonatal Ultra-Low-Field MRI}

\name{%
\parbox{\linewidth}{\centering
Malek Al Abed$^{1,3}$\thanks{M.~Al Abed and S.~Demir shared the first author position.},
Sebiha Demir$^{2}$,
Anne Groteklaes$^{2}$,
Elodie Germani$^{4}$\thanks{E.~Germani performed most of the work during her postdoc at the Albarqouni Lab, University Hospital Bonn.},\\[2pt]
Shahrooz Faghihroohi$^{3}$,
Hemmen Sabir$^{2}$,
Shadi Albarqouni$^{1,3}$\thanks{Corresponding author: shadi.albarqouni@ukbonn.de}
}}

 \address{$^{1}$ University of Bonn, University Hospital Bonn, Clinic for Diagnostic and Interventional Radiology, Germany \\
 $^{2}$ University of Bonn, University Hospital Bonn, Department of Experimental Neonatology, Germany\\
 $^{3}$ Technical University of Munich, School of Computation, Information and Technology, Germany\\
 $^{4}$ Laboratoire Traitement du Signal et de l’Image, Inserm UMR 1099, Université de Rennes, France\\
 }

\begin{document}
\ninept
\maketitle

\begin{abstract}

Portable ultra-low-field MRI (uLF-MRI, $0.064$\,T) enables accessible neonatal neuroimaging but suffers from low signal-to-noise ratio and limited diagnostic quality compared to high-field (HF) MRI. We propose \textbf{MRIQT}, a 3D conditional diffusion framework for \emph{image quality transfer} (IQT) from uLF to HF MRI. MRIQT combines physics-aware k-space degradation for realistic uLF simulation, $v$-prediction with classifier-free guidance for stable image-to-image generation, and an SNR-weighted 3D perceptual loss to preserve anatomical fidelity. MRIQT denoises from a noised uLF input conditioned on the same scan using a volumetric attention U-Net, enabling structure-preserving translation.
Trained on a neonatal cohort with diverse pathologies, MRIQT outperforms recent GAN- and CNN-based baselines in image fidelity and improves downstream anatomical consistency as measured by tissue segmentation Dice against HF references (\emph{PSNR} +$1.8\%$, \emph{Pearson's correlation} +$5.9\%$ \emph{Dice} +$9.4\%$ over the best baselines). In a reader study, clinicians rated $85\%$ of MRIQT reconstructions as good quality with clearly visible pathologies. Overall, MRIQT enables high-fidelity diffusion-based enhancement of portable uLF MRI for reliable neonatal brain assessment. Implementation is available online\footnote{\url{https://github.com/albarqounilab/MRIQT}}.

\end{abstract}

\begin{keywords}
Portable Ultra-Low-Field MRI, Image Quality Transfer, 3D-Diffusion Models.
\end{keywords}

\section{Introduction}
\label{sec:intro}

Monitoring neonatal brain development is crucial for assessing early neurological function and detecting developmental disorders, particularly since the brain is highly vulnerable to injury~\cite{groteklaes4888105identifying}. Neonatal brain injuries and disorders can lead to death, over 2.3 million deaths within the first few days of life~\cite{who2024newborn}, or long-term cognitive and motor impairments. Early detection of such pathologies is therefore essential to enable timely medical intervention, reduce infant mortality, and improve neuro-developmental outcomes. Neonatal neuroimaging aids in understanding that development.

In clinical practice, diagnosis typically begins with physical and neurological examinations, followed by \textit{cranial ultrasound (cUS)} for infants exhibiting abnormal signs. While cUS is fast and bedside-accessible, it is highly operator-dependent and prone to artifacts, often yielding inconclusive results. \textit{High-field magnetic resonance imaging (HF-MRI)} remains the gold standard for structural and functional assessment, providing high-resolution and high-contrast images of the developing brain. However, HF-MRI’s use in neonatal settings is severely constrained by the need for infant sedation, acoustic noise, prolonged scan times, limited portability, and high installation and maintenance costs, making it inaccessible in many neonatal intensive care units (NICUs).

\begin{figure}[!t]
  \centering
  \includegraphics[width=\linewidth]{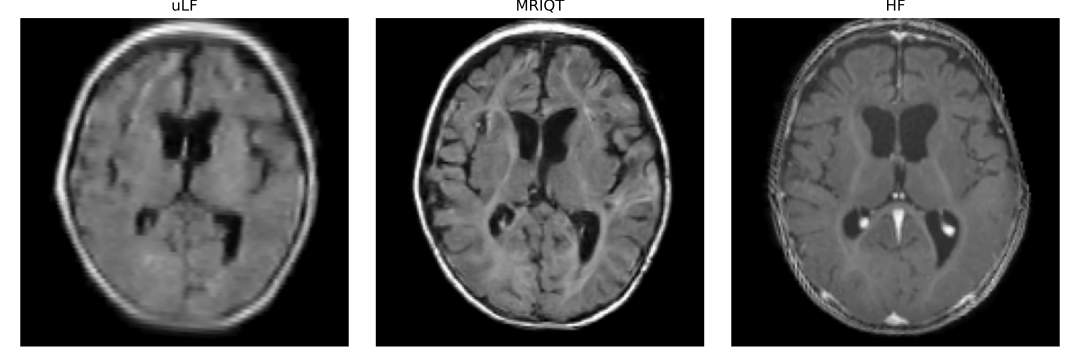}
\caption{Our MRIQT restores fine anatomical structures and contrast in portable uLF scans, producing HF-like images.}
\label{fig:teaser}
\end{figure}

Recently, \textit{ultra low-field MRI (uLF-MRI)} has emerged as a portable, low-cost, and bedside-compatible alternative. It significantly reduces scan time and obviates the need for sedation. Yet, the dramatic reduction in magnetic field strength yields a much weaker signal-to-noise ratio, poor tissue contrast, and reduced spatial resolution, thereby limiting its diagnostic reliability. Bridging the quality gap between uLF and HF MRI, known as \textbf{Image Quality Transfer (IQT)}, is thus crucial to unlocking the clinical potential of portable MRI systems for neonatal care. Enhancing such scans could facilitate the assessment process and aid in guiding intensive care therapy~\cite{sabir2023feasibility}.

Deep learning-based super-resolution (SR) and image-to-image translation methods have shown promise for IQT by learning mappings from low- to high-quality MRI domains. To date, most uLF-to-HF IQT efforts have been GAN-based, such as \textit{LoHiResGAN}~\cite{Islam2023improving}, which employs a slice-wise PatchGAN discriminator~\cite{demir2018patch}, and \textit{SFNet}~\cite{Tap_SuperField_MICCAI2024}, which integrates a Swin Transformer~\cite{liu2021swin} with dual-channel diffusion for uLF–HF data augmentation. Other approaches, including \textit{LF-SynthSR} from the Freesurfer suite~\cite{iglesias2021joint,
iglesias2023synthsr,iglesias2022quantitative}, use convolutional networks for generating isotropic T1-weighted scans, while \textit{GAMBAS}~\cite{baljer2025gambas} combines ConvNets with structured state-space blocks~\cite{gu2021efficiently} and Hilbert-curve serialization to model long-range spatial dependencies. Despite their effectiveness, GAN-based models often suffer from training instability, mode collapse, and poor structural fidelity, which are especially critical for clinical adoption. Moreover, most existing IQT models are trained on healthy adult or pediatric data, limiting their generalizability to neonatal pathologies.

Diffusion models, now the state-of-the-art in generative modeling, have recently demonstrated superior stability and fidelity in medical image synthesis. Prior works such as \textit{Med-DDPM}~\cite{mobaidoctor2023medddpm}, \textit{MU-Diff}~\cite{dayarathna2025mu}, and \textit{DiffusionIQT} $<1$T~\cite{kim20233d} explored MRI generation and super-resolution, yet these methods typically rely on simulated low-field inputs generated via simple downsampling and blurring of HF images, which fail to capture the true physics of low-field signal formation. Furthermore, none have specifically addressed neonatal populations with real pathological variability.

In this work, we introduce \textbf{MRIQT}, a fully 3D conditional diffusion framework for ultra-low- to high-field brain MRI quality transfer. Our main contributions are as follows:
\begin{enumerate}
    \item We propose a \textbf{fully 3D conditional diffusion model} for volumetric IQT between uLF and HF MRI, designed to preserve anatomical structures while enhancing resolution and contrast.
    \item We train \textbf{MRIQT} on a \textbf{neonatal-specific dataset} encompassing diverse pathologies (e.g., hemorrhage, tumors, and structural malformations), enabling IQT for real-world NICU deployment.
    \item We derive a \textbf{physics-aware K-space transfer function} from paired uLF–HF scans to generate realistic synthetic uLF data from HF images, which enables unpaired training yet generalization to real uLF acquisitions without paired supervision.
    \item We introduce a \textbf{3D VGG-like perceptual loss} that leverages feature-space similarity to guide diffusion sampling and early-sampling criteria, improving fine structural detail and convergence stability.
\end{enumerate}

Together, these innovations make \textbf{MRIQT} the first diffusion-based framework tailored for \textbf{clinically meaningful neonatal MRI quality transfer}, bridging the fidelity gap between portable uLF and diagnostic HF scans, and moving toward practical, accessible neuroimaging for early-life care.

\begin{figure*}[t]
  \centering
  \begin{subfigure}{0.4\textwidth}
    \centering
    \includegraphics[width=\linewidth]{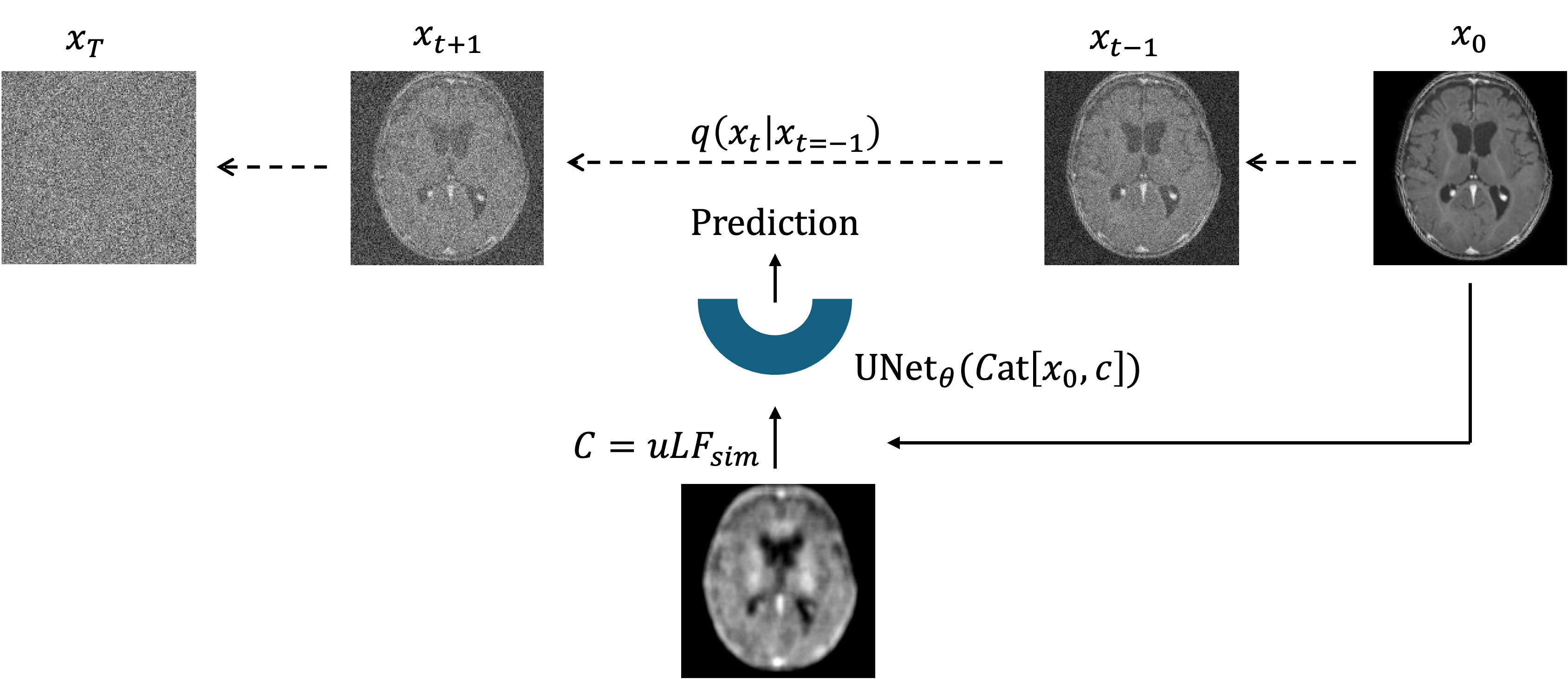}
    \caption{Training pipeline.}
    \label{fig:training}
  \end{subfigure}
  \hfill
  \begin{subfigure}{0.5\textwidth}
    \centering
    \includegraphics[width=\linewidth]{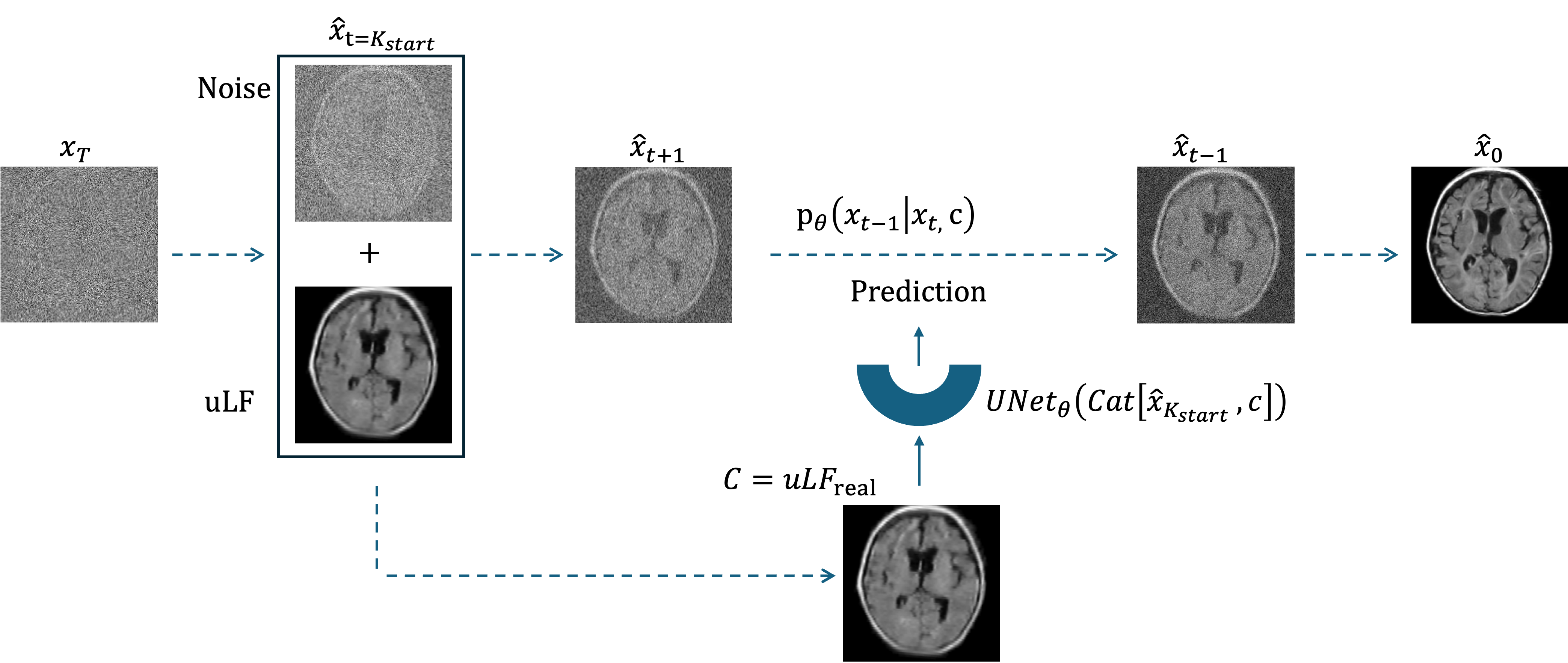}
    \caption{Inference pipeline.}
    \label{fig:inference_sub}
  \end{subfigure}

  \caption{Overview of the MRIQT framework showing the training (left) and inference (right) stages. Diffusion process from $x_0$ to $x_T$, where $x_0$ is HF reference and $x_t$ is noise ($\xleftarrow{}$). At each timestep $t \in T$, the UNet, conditioned on $c=\text{uLF}_{\text{sim}}$ is trained to predict the added noise/v. Inference/sampling starts from $\hat{x}_{t=K_{start}}=c+\epsilon$, where $c=\text{uLF}_{\text{real}}$, progressively denoising the input until $\hat{x_0}$.}
  \label{fig:training_inference}
\end{figure*}

\section{Methodology}
\label{sec:method}

We propose \textbf{MRIQT}, a 3D conditional diffusion framework for \textit{ultra-low-field to high-field (uLF–HF)} MRI quality transfer. MRIQT combines (i) a physics-based K-space degradation model for realistic uLF simulation, (ii) a 3D conditional DDPM with classifier-free guidance (CFG) and \(v\)-parametrization~\cite{salimans2022progressive}, and (iii) a 3D perceptual feature loss for structure-preserving generation and adaptive early stopping.

\subsection{Problem Setup}
Given uLF and HF volumes \(x_{\mathrm{uLF}}, x_{\mathrm{HF}} \in \mathbb{R}^{H\times W\times D}\), the goal is to learn a mapping \(\mathcal{F}_\theta: x_{\mathrm{uLF}} \!\rightarrow\! \hat{x}_{\mathrm{HF}}\) that preserves anatomy from uLF while recovering HF-level resolution and contrast. Training uses both real and simulated uLF–HF pairs, where the latter are generated from HF data using a learned K-space transfer function.

\subsection{K-space Simulation}
For paired samples, we compute Fourier transforms \(X_i(\mathbf{f})=\mathcal{F}\{x_i^{\mathrm{uLF}}\}\) and \(Y_i(\mathbf{f})=\mathcal{F}\{x_i^{\mathrm{HF}}\}\) and estimate a complex transfer function \(\hat{S}(\mathbf{f})\) by solving a Tikhonov-regularized least squares:
\[
\hat{S}(\mathbf{f})=\arg\min_{S}\sum_i|X_i(\mathbf{f})-S(\mathbf{f})Y_i(\mathbf{f})|^2+\lambda|S(\mathbf{f})|^2.
\]
Synthetic uLF data are generated as \(\tilde{x}_{\mathrm{uLF}}=\mathcal{F}^{-1}(\hat{S}(\mathbf{f})\!\cdot\! Y(\mathbf{f}))\). Matching radial power spectra confirms that simulated and real uLF scans share similar frequency behavior, enabling unpaired training on large HF datasets.

\subsection{3D Conditional Diffusion}
MRIQT extends the DDPM~\cite{ho2020denoising} to 3D volumes. The forward process adds Gaussian noise \(x_t=\sqrt{\bar{\alpha}_t}x_0+\sqrt{1-\bar{\alpha}_t}\epsilon\) (Figure ~\ref{fig:training}), while the reverse model learns to predict and remove the noise conditioned on the uLF input. Instead of unconditional sampling, we use \textbf{image-to-image diffusion}, initializing denoising from a partially noised uLF scan, which preserves structural fidelity while allowing generative refinement (Figure~\ref {fig:inference_sub}), as in~\cite{saharia2022image}. 

\subsection{Classifier-Free Guidance}
To balance uLF faithfulness and HF detail, we apply CFG~\cite{ho2022classifier}. During training, conditional and unconditional modes are alternated. At inference, predictions are merged as 
\(\hat{y}_{\mathrm{cfg}}=\hat{y}_{\mathrm{uc}}+w(\hat{y}_{\mathrm{c}}-\hat{y}_{\mathrm{uc}})\), 
where \(w\) adjusts the strength of guidance and \(\hat{y}\in\{\hat{\epsilon},\hat{v}\}\).

\subsection{\(v\)-Parametrization}
Instead of predicting noise \(\epsilon\), we use the stable \(v\)-prediction~\cite{salimans2022progressive}, where \(v=\sqrt{\bar{\alpha}_t}\epsilon-\sqrt{1-\bar{\alpha}_t}x_0\). The model predicts \(\hat{v}_\theta([x_t,c],t)\), from which we recover \(\hat{x}_0=\sqrt{\bar{\alpha}_t}x_t-\sqrt{1-\bar{\alpha}_t}\hat{v}_\theta\). This yields smoother gradients across timesteps and sharper structural details than $\epsilon$-prediction.

\subsection{3D Perceptual Feature Extractor}
Conventional 2D perceptual networks in MONAI~\cite{zhang2018unreasonable} fail on volumetric data. We therefore train a \textbf{3D VGG-like feature extractor} on our neonatal HF dataset. The perceptual loss both (i) enhances contrast and fine texture during training and (ii) identifies an optimal sampling start step \(K<T\) where uLF and HF become perceptually similar, improving efficiency and convergence.

\subsection{Training Objective}
The total loss combines \(v\)-prediction with SNR-weighted perceptual alignment:
\[
\mathcal{L}_{\text{total}}
= \|v_{\text{pred}}-v_{\text{true}}\|_2^2
+ \lambda_{\text{p}}\,w_{\text{SNR}}(t)
\sum_{l\in L}w_l\,\mathrm{SmoothL1}(\phi_l(\hat{x}_0),\phi_l(x_0)),
\]
where \(\phi_l\) denotes multi-scale feature maps from the 3D extractor, \(w_l\) are layer weights, \(w_{\text{SNR}}(t)\) down-weights perceptual loss at high-noise steps, and $\lambda_{\text{p}}$ is a regularization parameter.

\section{Experiments and Results}
\label{sec:exps}

\begin{table*}[htbp]
\centering
\caption{
Quantitative comparison on the \textbf{paired subset} (34 LF–HF pairs with matched T2w for \cite{iglesias2022quantitative}).  
indications $\uparrow$ higher is better; $\downarrow$ lower is better.  
Bold indicates the best result in each column. \underline{underlined} values are a close second. 
Significance markers: $(*)$ significant ($p<0.01$). 
}
\label{tab:results_subset}
\setlength{\tabcolsep}{4pt}
\renewcommand{\arraystretch}{1.15}
\resizebox{\textwidth}{!}{%
\begin{tabular}{lccccccc}
\hline
\textbf{Method} & \textbf{PSNR}$\uparrow$ & \textbf{Pearson}$\uparrow$ & \textbf{SSIM}$\uparrow$ & \textbf{MS-SSIM}$\uparrow$ & \textbf{LPIPS}$\downarrow$ & \textbf{MAE}$\downarrow$ & \textbf{DICE}$\uparrow$ \\
\hline
Baseline [uLF-HF] & ${15. 84 \pm 0.870\,*}$ & $0.388 \pm 0.071\,*$ & ${0.587 \pm 0.029\,*}$ & $ {0.637 \pm 0.029\,*}$ & $0.318 \pm 0.059\,*$ & ${0.183 \pm 0.027\,*}$ & ${0.481 \pm 0.054}$ \\
\hline
LoHiResGAN \cite{Islam2023improving} & \underline{$15.07 \pm 0.920$} & $0.390 \pm 0.066$ & $0.560 \pm 0.020$ & $0.612 \pm 0.022$ & $0.290 \pm 0.023\,*$ & $\mathbf{0.185 \pm 0.035}$ & ${0.447 \pm 0.049*}$ \\
LF-SynthSR \cite{iglesias2021joint,iglesias2022quantitative} & $10.45 \pm 1.265\,*$ & $0.214 \pm 0.054\,*$ & $0.492 \pm 0.017\,*$ & $0.510 \pm 0.023\,*$ & $0.317 \pm 0.024\,*$ & $0.415 \pm 0.063\,*$ & ${0.438 \pm 0.043*}$ \\
SFNet \cite{Tap_SuperField_MICCAI2024} & $14.90 \pm 1.371$ & $0.345 \pm 0.068\,*$ & $\mathbf{0.575 \pm 0.028\,*}$ & $\mathbf{0.626 \pm 0.028}$ & $\mathbf{0.219 \pm 0.020\,*}$ & $0.228 \pm 0.047\,*$ & ${0.369 \pm 0.080*}$ \\
GAMBAS \cite{baljer2025gambas} 
& $14.05 \pm 1.278\,*$ & $0.314 \pm 0.082\,*$ & $0.524 \pm 0.018\,*$ & $0.540 \pm 0.023\,*$ & $0.287 \pm 0.031\,*$ & $0.228 \pm 0.042\,*$ & ${0.391 \pm 0.055*}$ \\
\hline
\textbf{MRIQT} &  &  &  &  &  & \\
\quad $\epsilon$-pred, $\lambda_{p}=0.25$ & $15.23 \pm 1.006\,*$ & $\underline{0.410 \pm 0.072}$ & \underline{$0.569 \pm 0.031\,*$} & \underline{$0.622 \pm 0.033\,*$} & $0.277 \pm 0.028\,*$ & $0.195 \pm 0.036$ & $\mathbf{0.488 \pm 0.048*}$ \\

\quad $v$-pred, $\lambda_{p}=0$ & $15.18 \pm 1.077\,*$ & $\mathbf{0.413 \pm 0.074\,*}$ & $0.567 \pm 0.031\,*$ & $0.621 \pm 0.033\,*$ & $0.252 \pm 0.028\,*$ & $0.198 \pm 0.040\,*$ & ${\mathbf{0.489 \pm 0.049*}}$ \\

\quad $v$-pred, $\lambda_{p}=0.25$ (Ours) & $\mathbf{15.34 \pm 1.046}$ & $0.407 \pm 0.073$ & $0.561 \pm 0.027$ & $0.617 \pm 0.031$ & \underline{$0.239 \pm 0.023$} & \underline{$0.194 \pm 0.038$} & $\underline{{0.474 \pm 0.043*}}$ \\

\hline
\end{tabular}
}%
\end{table*}

\begin{figure*}[t]
  \centering 
  \includegraphics[width=\linewidth]{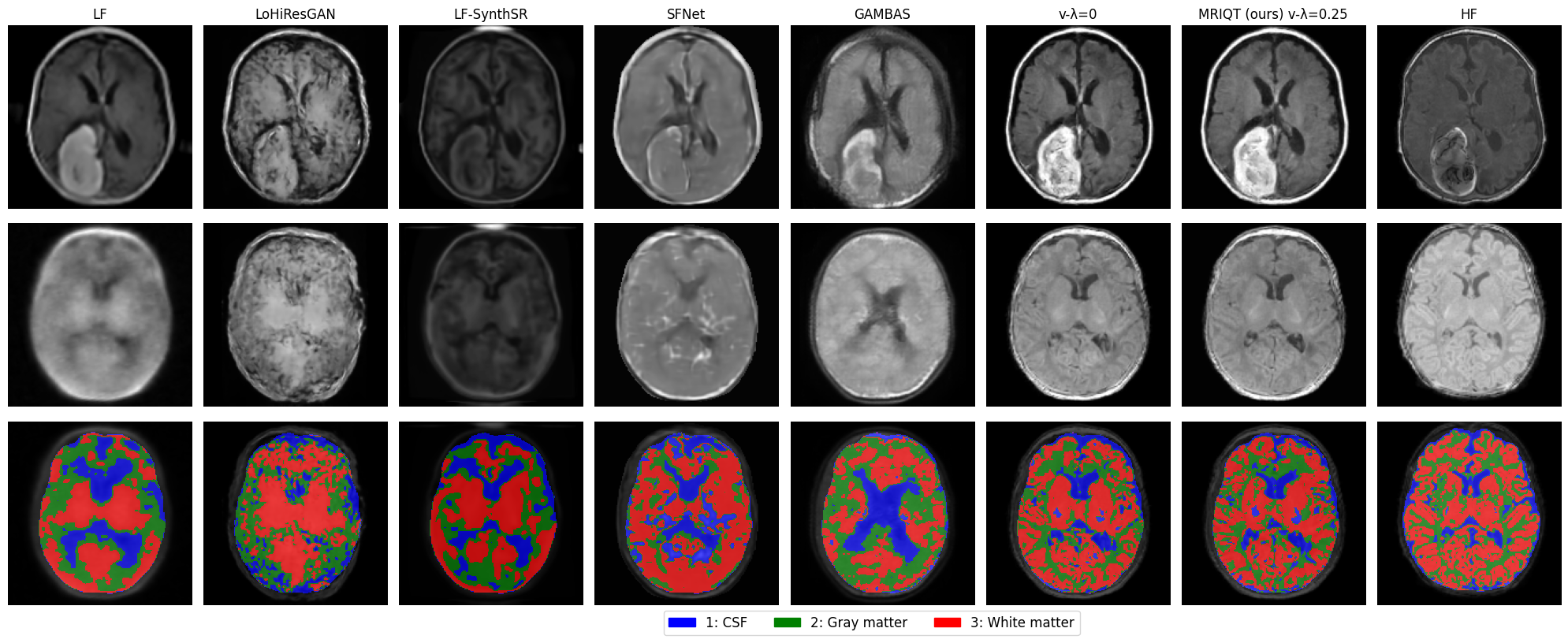}
  \caption{Qualitative comparison of 2 samples on the axial view. The first and second rows correspond to the generated samples, while the third row presents the corresponding tissue segmentation results for the second sample. Left to right: ULF, LoHiResGAN \cite{Islam2023improving}, LF-SynthSR$\ddagger$ \cite{iglesias2022quantitative}, SFNet$\dagger$ \cite{Tap_SuperField_MICCAI2024}, GAMBAS$\dagger$ \cite{baljer2025gambas}, Our base model, Ours, reference HF. [($\dagger$) trained on T2w-scans, ($\ddagger$) testing requires both T1w and T2w scans.]}
  \label{fig:results_}
\end{figure*}

We evaluate \textbf{MRIQT} across quantitative, qualitative, and ablation studies to assess its ability to enhance uLF brain MRI quality and fidelity toward HF references. Experiments are designed to (i) validate the accuracy of our physics-based uLF simulation, (ii) benchmark MRIQT against state-of-the-art IQT and SR models, and (iii) isolate the contribution of key design components, including $v$-prediction, perceptual loss, and CFG weighting.

\subsection{Experimental Setup}
\subsubsection{Dataset}
All data were collected at the Department of Neonatology and Pediatric Intensive Care Medicine, University Hospital Bonn, using both HF and uLF MRI scanners. T1-weighted uLF data were acquired using a 64\,mT SWOOP (Hyperfine) scanner (1.5$\times$1.5$\times$2\,mm$^3$ anisotropic resolution). HF data were obtained from multiple clinical 3\,T scanners with heterogeneous resolutions ranging from (0.35$\times$0.34$\times$0.43\,mm$^3$) to (0.89$\times$ 0.89 $\times$2\,mm$^3$). The cohort included neonates under six months of age with nine different pathologies. After quality control, the dataset comprised 50 matched uLF–HF pairs, 100 HF-only scans, and 130 uLF-only scans. Matched pairs were used for K-space transfer estimation, $K$ determination, and final testing, while unpaired HF data were used for training and unpaired uLF data for reader study evaluation.
All scans were converted from DICOM to NIfTI and organized following the BIDS standard~\cite{gorgolewski2016brain}. Preprocessing used ANTs~\cite{
avants2011reproducible}: bias-field correction, resampling to 1\,mm$^3$, affine registration to the MNI152NLin2009cAsym template for 0–2 months~\cite{fonov2009unbiased}, and cropping to a $(160,160,160)$ voxel grid. 

\subsubsection{Implementation Details}
All experiments were conducted on an NVIDIA A100 (80\,GB) GPU using PyTorch, with our implementation built upon the Med-DDPM framework~\cite{mobaidoctor2023medddpm}.  
Our 3D residual U-Net with attention bottleneck was trained for 30{,}000 epochs using AdamW with a $2\!\times\!10^{-5}$ weight decay, a 1{,}500-epoch warmup, and a cosine annealing scheduler (peak LR: $2\!\times\!10^{-5}$). Diffusion steps were $T{=}1000$; inference using the learned step \textbf{$K$ achieved a 35\% speedup} over standard sampling, from \emph{18.5 to 11 minutes}. The best CFG setup used $cond\_drop\_prob{=}0.1$ and $guidance\_weight{=}2$. The perceptual weight was fixed at $\lambda_{\text{p}}{=}0.25$.  
The 3D VGG-like feature extractor was trained via 5-fold cross-validation ($K_{\text{cv}}{=}5$) using Adam and a linearly decaying LR schedule. 

\subsubsection{Baselines and Evaluation Metrics}
We compare MRIQT against four state-of-the-art (SoTA) IQT and SR methods: \textbf{LoHiResGAN}~\cite{Islam2023improving}, \textbf{LF-SynthSR}~\cite{iglesias2021joint,iglesias2022quantitative}, \textbf{SFNet}~\cite{Tap_SuperField_MICCAI2024}, and \textbf{GAMBAS}~\cite{baljer2025gambas}. Tests were conducted on 46 matched uLF–HF pairs (34 for LF-SynthSR, which requires T2-weighted input). Note that SFNet and GAMBAS were pre-trained on T2-weighted data, whereas MRIQT focuses on T1-weighted MRI.

We employed complementary evaluation metrics;
\emph{Pixel-level metrics}: PSNR, MAE, and Pearson correlation to quantify reconstruction accuracy. \emph{perceptual and structural metrics}: (MS-)SSIM, and LPIPS to assess perceptual realism and structural fidelity.

In addition, for downstream anatomical consistency, we ran autosegmentation with FAST~\cite{zhang2002segmentation}, computing Dice scores for cerebrospinal fluid (CSF), gray matter (GM), and white matter (WM) against the reference HF scans.
Statistical significance \emph{w.r.t.} \textbf{Ours} was assessed using paired t-tests on the paired test set at $p < 0.01$.

\subsection{Quantitative Comparison}
Table~\ref{tab:results_subset} summarizes results on the paired subset (34 T1w samples with matched T2w scans). Conventional fidelity metrics—PSNR, MAE, and (MS-)SSIM—primarily capture voxel-wise and structural intensity agreement rather than perceptual realism or anatomical plausibility, as illustrated by the baseline results, where high PSNR and SSIM do not correspond to superior visual quality.

MRIQT achieves the highest PSNR ($15.34$\,dB) and Pearson correlation ($0.413$), with competitive (MS-)SSIM, indicating effective noise suppression and detail recovery. While SFNet attains slightly higher SSIM and lower LPIPS due to strong perceptual weighting, it can exaggerate local details. In contrast, MRIQT yields lower MAE than GAN-based methods, supporting anatomically consistent denoising without over-smoothing. None of the competing methods are statistically superior to MRIQT in these metrics.

Downstream tissue segmentation further validates anatomical consistency against the HF reference. MRIQT achieves the highest Dice scores across all tissues, whereas methods optimizing single metrics exhibit reduced segmentation accuracy. Overall, MRIQT provides a balanced trade-off between fidelity and perceptual quality while achieving the highest segmentation scores.

\subsection{Ablation Studies}

Replacing $v$-prediction with $\epsilon$-prediction decreases PSNR and LPIPS, confirming $v$-prediction’s stability and fine-detail advantage. Removing the perceptual loss ($\lambda_{\text{p}}{=}0$) leads to minor degradation in structure metrics (SSIM, MS-SSIM), showing that perceptual supervision primarily enhances contrast and local sharpness. The CFG hyperparameter sweep indicated that moderate guidance ($w{=}2$) balances realism and fidelity, avoiding hallucinated features. Overall, each contribution, $v$-prediction, perceptual loss, and CFG, synergistically improves quantitative and perceptual outcomes.

\subsection{Qualitative Analysis}

Figure~\ref{fig:results_} compares uLF, SoTA, and MRIQT reconstructions across two representative samples (first 2 rows). Competing GAN-based models (e.g., LoHiResGAN, SFNet) tend to oversharpen or hallucinate fine structures, whereas MRIQT consistently restores cortical boundaries and deep gray matter contrast without introducing artifacts. 
MRIQT also preserves pathological features such as hemorrhagic lesions and ventricular enlargement, closely aligning with the HF reference. The segmentation results on the second sample (third row) further highlight MRIQT’s superior anatomical consistency, showing clearer tissue separation and reduced misclassification compared to competing methods. Overall, the denoising smoothness and texture recovery confirm that diffusion-based IQT generalizes effectively across unseen pathologies.

\subsection{Reader Study}
Three experienced neonatal neuroimaging raters independently evaluated $34$ generated scans, scoring image quality as bad, acceptable, or good ($0$, $1$, $2$). Pathology presence was assessed as a quality indicator. We achieved a mean pathology detection accuracy of $80\%$ and sensitivity of $82.3\%$, with an average image quality score of $1.6$. Raters were not shown the original scans; therefore, the study is not clinically applicable, and diagnostic decisions were not permitted due to ethical constraints.

\section{Conclusion}
\label{sec:conclusion}

We presented \textbf{MRIQT}, a 3D conditional diffusion framework for uLF-to-HF brain MRI image quality transfer. Tailored to neonatal imaging, MRIQT integrates physics-aware k-space simulation, $v$-prediction, classifier-free guidance, and a 3D perceptual loss to enhance uLF scans while preserving anatomical fidelity.

Experiments on neonatal data demonstrate that MRIQT outperforms GAN- and CNN-based baselines in PSNR, perceptual quality, and downstream autosegmentation Dice, without introducing hallucinations. Reader studies and segmentation results further support the interpretability and diagnostic plausibility of the reconstructions.

Future work will extend to multimodal, real-time reconstruction, enabling high-quality bedside neuroimaging in NICUs and resource-limited areas, as well as pathology-specific generation and analysis with a diverse transfer function ${S}$ and further uncertainty quantification for reliable outcomes. 

\clearpage

\noindent \textbf{Compliance with ethical standards.} This study was performed in line with the principles of the Declaration of Helsinki. Approval was granted by the Ethics Committee of UniBonn (Ethics Nr. 167/22).

\bibliographystyle{IEEEbib}
\bibliography{strings,refs}

\end{document}